\pdfoutput=1

\documentclass[11pt]{article}

\usepackage{emnlp2021}

\usepackage{times}
\usepackage{latexsym}
\usepackage{microtype}
\usepackage{multirow}
\usepackage{colortbl}
\usepackage{booktabs}
\usepackage{amssymb}
\usepackage{amsmath}
\usepackage{amsthm}
\usepackage{graphicx}
\usepackage[T1]{fontenc}

\usepackage[utf8]{inputenc}

\usepackage{microtype}
\title{Thinking Clearly, Talking Fast: Concept-Guided Non-Autoregressive Generation for Open-Domain Dialogue Systems}

\author{Yicheng Zou,\ \ Zhihua Liu,\ \ Xingwu Hu,\ \  Qi Zhang\\
  Shanghai Key Laboratory of Intelligent Information Processing, Fudan University\\
  School of Computer Science, Fudan University\\
  Shanghai, China\\
  \texttt{\{yczou18,liuzh20,xwhu20,qz\}@fudan.edu.cn}}

\begin{document}
\maketitle
\begin{abstract}

Human dialogue contains evolving concepts, and speakers naturally associate multiple concepts to compose a response. However, current dialogue models with the seq2seq framework lack the ability to effectively manage concept transitions and can hardly introduce multiple concepts to responses in a sequential decoding manner. To facilitate a controllable and coherent dialogue, in this work, we devise a concept-guided non-autoregressive model (CG-nAR) for open-domain dialogue generation. The proposed model comprises a multi-concept planning module that learns to identify multiple associated concepts from a concept graph and a customized Insertion Transformer that performs concept-guided non-autoregressive generation to complete a response. The experimental results on two public datasets show that CG-nAR can produce diverse and coherent responses, outperforming state-of-the-art baselines in both automatic and human evaluations with substantially faster inference speed.

\end{abstract}

\section{Introduction}

Creating a "human-like" dialogue system is one of the important goals of artificial intelligence. Recently, due to the rapid advancements in natural language generation (NLG) techniques, data-driven approaches have attracted lots of research interest and have achieved impressive progress in producing fluent dialogue responses \cite{shang2015neural,vinyals2015neural,serban2016building,li2016diversity}. However, such seq2seq models tend to degenerate generic or off-topic responses \cite{tang2019target,welleck2019neural}. An effective way to address this issue is to leverage external knowledge \cite{zhou2018commonsense,zhou2018dataset} or topic information \cite{xing2017topic}, which are integrated as additional semantic representations to improve dialogue informativeness. 

Although promising results have been obtained by equipping dialogue models with external knowledge, the development of dialogue discourse still has its own challenge: human dialogue generally evolves around a number of concepts that might frequently shift in a dialogue flow \cite{zhang2020grounded}. The lack of concept management strategies might lead to incoherent dialogue due to the loosely connected concepts. To address this problem, recent studies have combined concept planning with response generation to form a more coherent and controllable dialogue \cite{wu2019proactive,xu2020conversational,xu2020knowledge,wu2020topicka,zhang2020grounded}.

\begin{figure}
\centering
  \includegraphics[width=2.8in]{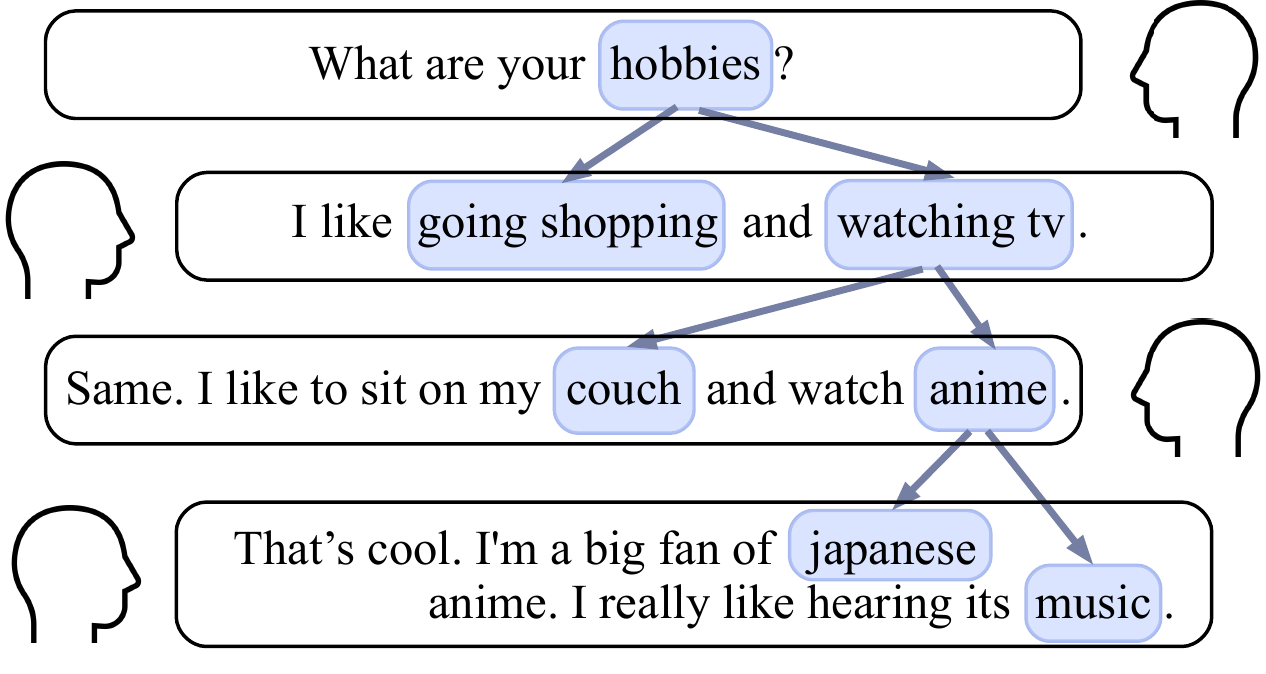}
  \caption{An exemplar dialogue with concept transitions, where each utterance is composed of multiple associated concepts to convey diverse information.} \label{fig:intro}
\end{figure}

Most of these approaches incorporate concepts into responses in an implicit manner, which cannot guarantee the appearance of a concept in a response. Compared with dialogue concepts, a large proportion of chit-chat words are common and usually have a high word frequency and are relatively over-optimized in language models \cite{gong2018frage,khassanov2019enriching}. Consequently, conventional seq2seq generators are more "familiar" with these generic words than those requiring concept management, which prevents introducing certain concepts to the response with sequential decoding (either greedily or with beam search) \cite{mou2016sequence}. Moreover, speakers naturally associate multiple concepts to proactively convey diverse information, e.g., action, entity, and emotion (see Figure \ref{fig:intro}). Unfortunately, most existing methods can only retrieve one concept for each utterance \cite{tang2019target,qin2020dynamic}. Another line of approaches attempt to explicitly integrate concepts into responses and generate the remaining words in both directions \cite{mou2016sequence,xu2020conversational}, but they also fail to deal with multiple concepts.

In this paper, we devise a concept-guided non-autoregressive model (CG-nAR) to facilitate dialogue coherence by explicitly introducing multiple concepts into dialogue responses. Specifically, following Xu et al. \shortcite{xu2020conversational}, a concept graph is constructed based on the dialogue data, where the vertices represent concepts, and edges represent concept transitions between utterances. Based on the concept graph, we introduce a novel multi-concept planning module that learns to manage concept transitions in a dialogue flow. It recurrently reads historical concepts and dialogue context to attentively select multiple concepts in the proper order, which reflects the transition and arrangement of target concepts. Then, we customize an Insertion Transformer \cite{stern2019insertion} by initializing the selected concepts as a partial response for subsequent non-autoregressive generation. The remaining words of a response are generated in parallel, aiming to foster a fast and controllable decoding process. 

We conducted experiments on Persona-Chat \cite{zhang2018personalizing} and Weibo \cite{shang2015neural}. The results of automatic and human evaluations show that CG-nAR achieves better performance in terms of response diversity and dialogue coherence. We also show that the inference time of our model is much faster than conventional seq2seq models. All our codes and datasets are publicly available.\footnote{\url{https://github.com/RowitZou/CG-nAR}}

Our contributions to the field are three-fold: 1) We design a concept-guided non-autoregressive strategy that can successfully integrate multiple concepts into responses for a controllable decoding process. 2) The proposed multi-concept planning module effectively manages multi-concept transitions and remedies the problem of dialogue incoherence. 3) Comprehensive studies on two datasets show the effectiveness of our method in terms of response quality and decoding efficiency. 

\section{Related Work}
\subsection{Open-Domain Dialogue Generation}
Neural seq2seq models \cite{sutskever2014sequence} have achieved remarkable success in dialogue systems \cite{shang2015neural,vinyals2015neural,serban2016building,xing2017topic}, but they prefer to produce generic and off-topic responses \cite{tang2019target,welleck2019neural}. Dozens of works have attempted to incorporate external knowledge into dialogue systems to improve informativeness and diversity \cite{zhou2018commonsense,zhang2018personalizing,dinan2018wizard,ren2020thinking}. Beyond the progress on response quality, a couple of works focus on goal planning or concept transition for a controllable and coherent dialogue \cite{yao2018chat,moon2019opendialkg,wu2019proactive,xu2020conversational,xu2020knowledge,wu2020topicka,zhang2020grounded}. Most of these works mainly explore how to effectively leverage external knowledge graphs and extract concepts from them. Nevertheless, they generally introduce concepts into the response implicitly with gated controlling or copy mechanism, which cannot ensure the success of concept integration because seq2seq models prefer generic words. Some works \cite{mou2016sequence,xu2020conversational} try to produce concept words first and generate the remaining words to both directions to complete a response, but they cannot handle the situation of multiple concepts. By contrast, we focus on how to effectively integrate multiple extracted concepts into dialogue responses. The proposed CG-nAR applies the non-autoregressive mechanism, which can explicitly introduce multiple concepts simultaneously to responses to enhance coherence and diversity.

\begin{figure*}
\centering
  \includegraphics[width=6.0in]{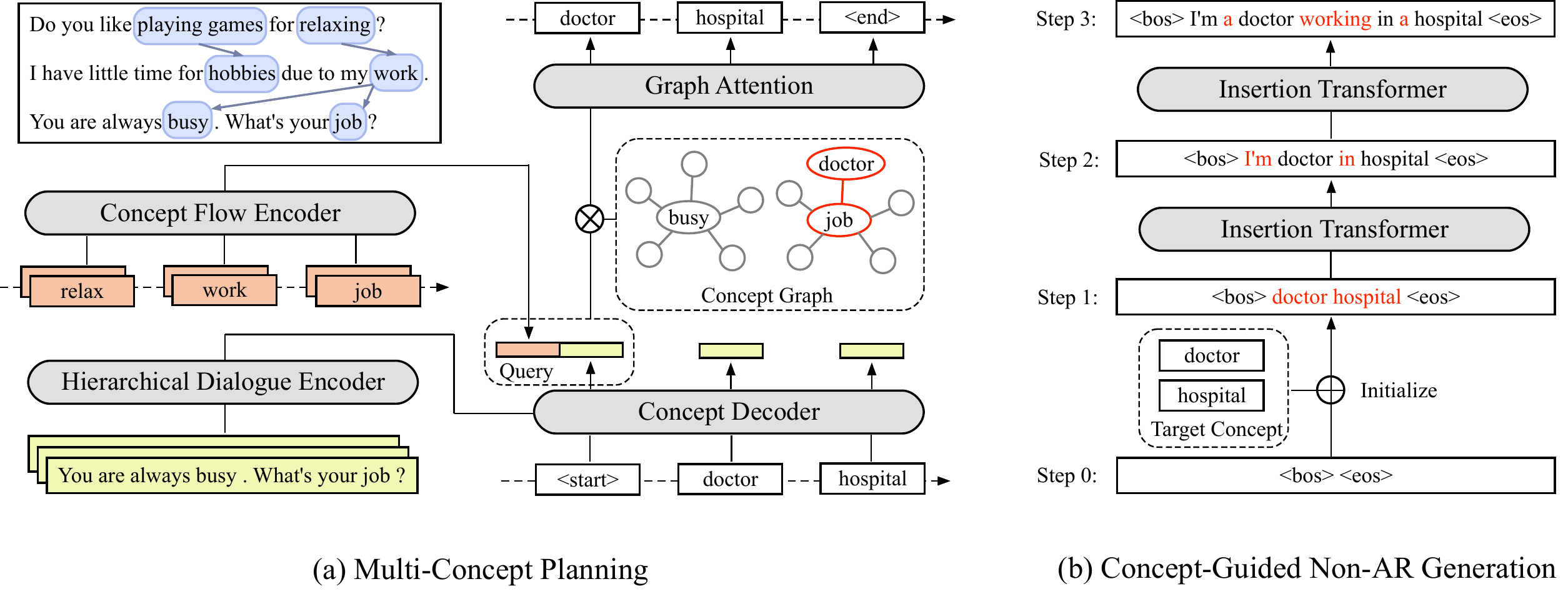}
  \caption{The overall framework of CG-nAR. (a) The multi-concept planning module conditions on the previous concept flow and the dialogue context to attentively select multiple associated concepts from the concept graph. (b) The selected concepts are used to initialize a partial response for subsequent non-autoregressive generation. } \label{fig:model}
\end{figure*}

\subsection{Non-Autoregressive Generation}

Compared with traditional sequential generators that conditions each output word on previously generated outputs, non-autoregressive (non-AR) generation avoids this property to speed up decoding efficiency and has recently attracted much attention \cite{gu2018nonautoregressive,gu2019insertion,ma2019flowseq,stern2019insertion}. Another relevant line of research is refinement-based generation \cite{lee2018deterministic,kasai2020non,hua2020pair,tan2021progressive}, which gradually improves generation quality by iterative refinement on the draft instead of one-pass generation. For dialogue systems, there has been prior works that attempt to improve the traditional autoregressive generation. Mou et al. \shortcite{mou2016sequence} explores the way of generating words to both directions, but it is still in an autoregressive manner. 
Song et al. \shortcite{song2020generate} introduces a three-stage refinement strategy for improving persona consistency of dialogue generation, but it requires a specialized consistency matching model for inference. Han et al. \shortcite{han2020non} applies the non-AR mechanism to dialogue generation, aiming to alleviate the non-globally-optimal issue to produce a more diverse response. In this work, we further use dialogue concepts to guide response generation. We customize an Insertion Transformer \cite{stern2019insertion} and arrange dialogue concepts as a partial input sequence, which is different from the original setting where texts are generated from scratch. By this means, multiple concepts can be naturally introduced to the response as a guidance to foster a more controllable non-AR generation.

\section{Methodology}
The overall framework of CG-nAR is shown in Figure \ref{fig:model}. Based on a concept graph that represents candidate concept transitions, a multi-concept planning module is designed to select and arrange appropriate target concepts from the contextually related subgraphs, which is conditioned on the previous concept flow and the dialogue context. Then, we input the selected concepts as a partial response into an Insertion Transformer \cite{stern2019insertion} to parallelly generate the remaining words.

\subsection{Concept Graph Construction}

Inspired by Xu et al. \shortcite{xu2020conversational}, we build a concept graph with two steps: vertex construction\footnote{The original constructed vertices in Xu et al. \shortcite{xu2020conversational} involve what-vertices and how-vertices, where how-vertices represent different ways of expressing response content with a multi-mapping model \cite{chen2019generating}. Here, we only collect what-vertices as dialogue concepts.} and edge construction. Given a dialogue corpus $S$, we exploit a rule-based keyword extraction method to identify salient keywords from utterances in $S$ \cite{tang2019target}. All extracted keywords are collected as dialogue concepts that represent vertices in the concept graph. For edge construction, we use pointwise mutual information (PMI) \cite{church1990word} to construct a concept pairwise matrix that characterizes the association between concepts in the observed dialogue data \cite{mou2016sequence,tang2019target}, where each concept pair consists of two concepts that are extracted from the context and the response, respectively. For each head vertex $v^h$, we select concepts with top PMI scores as tail vertices $v^t$ and build edges by connecting $v^h$ with all $v^t$s. In this way, we filter out low-frequency edges to narrow the search space for downstream concept planning.

\subsection{Multi-Concept Planning Module}
Given the dialogue context $D$, the historical concept flow $F$, and a concept graph $\mathcal{G}$, the goal of multi-concept planning is to predict a sequence of target concepts $C$, namely $P(C|D,F,\mathcal{G})$. All target concepts are extracted from $\mathcal{G}$ and arranged in a sequence $C=\{c_1,c_2,...,c_t\}$, which reflects the order of target concepts in the final response. 

{\bf Hierarchical Dialogue Encoder.} To facilitate the understanding of dialogue context $D$, we employ Transformer blocks \cite{vaswani2017attention} to hierarchically encode dialogue context, aiming to capture the global semantic dependency between utterances. Formally, given the dialogue context $D=\{u_1,u_2,...,u_N\}$ with $N$ utterances, where $u_i=\{w_{i1}, w_{i2},...,w_{in}\}$ is the word sequence of $i$-th utterance, we transform $u_i$ into a sequence of hidden vectors with a Transformer encoder:
\begin{equation}
    [\mathbf{\hat{h}}_i^{cls},\mathbf{\hat{h}}_{i1},...,\mathbf{\hat{h}}_{in}] = \mathrm{TF}_{\theta_w}([\mathbf{e}_i^{cls}, \mathbf{e}_{i1}^w,..., \mathbf{e}_{in}^w]).
\end{equation}
Here, $\mathbf{e}_{ij}^w$ is the embedding of the $j$-th word in $u_i$. $\mathbf{\hat{h}}_i^{cls}$ and $\mathbf{e}_i^{cls}$ represent a special token $\mathrm{[CLS]}$ that is used to aggregate sequence representations, which is inspired by Devlin et al. \shortcite{devlin2019bert}. Then, we collect utterance representations derived from $\mathrm{[CLS]}$ and input them into another Transformer encoder to hierarchically fuse context information:
\begin{equation}
    [\mathbf{h}_1^{cls},\mathbf{h}_2^{cls},...,\mathbf{h}_N^{cls}] = \mathrm{TF}_{\theta_u}([\mathbf{\hat{h}}_1^{cls}, \mathbf{\hat{h}}_2^{cls},..., \mathbf{\hat{h}}_N^{cls}]).
\end{equation}
$\mathbf{h}_i^{cls}$ is a context-aware utterance representation that can be used to guide concept selection in the following steps. 

{\bf Concept Flow Encoder.} Formally, a concept flow $F=\{f_1,f_2,...,f_N\}$ represents the observed concepts in the dialogue context, where $f_i$ means a concept set corresponding to the $i$-th utterance that collects all the concept words in $u_i$, namely $f_i=\{c_{i1},c_{i2},...c_{im}\}$. Here, $c_{ij}$ is the $j$-th concept word in $u_i$. For an empty set $f_i=\emptyset$, a special NULL token is served as the concept word.

To capture information of history concept transitions, we exploit a vanilla GRU unit \cite{cho2014learning} to recursively read concept words in the flow:
\begin{equation}
    \mathbf{s}_i = \mathrm{GRU}(\mathbf{s}_{i-1}, \mathbf{f}_i),\quad i\in[1,N].
\end{equation}
Here $\mathbf{f}_i$ denotes the representation of concept set $f_i$, which is calculated as a weighted sum of concept word embeddings $\mathbf{e}^{c}_{ij}$:
\begin{align}
    \mathbf{f}_i &=\sum\nolimits^m_{j=1}\alpha^f_{ij}\mathbf{e}^{c}_{ij},\nonumber \\
    \alpha^f_{ij} &= \frac{\mathrm{exp}(\beta^f_{ij})}{\sum^m_{k=1}\mathrm{exp}(\beta^f_{ik})}, \\
    \beta^f_{ij} &= \mathbf{s}_{i-1}^{\top}\mathbf{W}_f \ \mathbf{e}_{ij}^{c},\nonumber
\end{align}
where $\mathbf{W}_f$ is trainable parameters. The output state $\mathbf{s}_{i-1}$ at the $i-1$ step is used as a query to compute $\beta^f_{ij}$ scores, which can measure the preference of transitions to associated concepts. Empirically, $\mathbf{s}_0$ is a zero vector to initialize the recurrent process, and the final output $\mathbf{s}_N$ can serve as a memory to enable history-aware concept planning.

{\bf Multi-Concept Extractor.} Recall that our goal is to produce a concept sequence $C$, which is a subsequence of the target response. Inspired by {\em pointer network} \cite{vinyals2015pointer}, we design a multi-concept extractor to achieve this goal, which can attentively read the dialogue context and the concept flow to sequentially extract target concepts from the contextually related subgraphs in $\mathcal{G}$.

To implement concept extraction in a sequential decoding manner, we use a Transformer decoder and compute its decoding states as follows: 
\begin{align}
    \mathbf{H}^{cls}=[\mathbf{h}^{cls}_1,\mathbf{h}^{cls}_2,...,\mathbf{h}^{cls}_N],\nonumber\\
    \mathbf{m}_t= \mathrm{TF}_{\theta_d}([\mathbf{e}^{c}_{1:t-1}],\mathbf{H}^{cls}).
\end{align}
The utterance representations $\mathbf{H}^{cls}$ are memories for decoder-encoder attention. $[\mathbf{e}^{c}_{1:t-1}]$ denotes the embeddings of previously decoded concepts. $\mathbf{m}_t$ is the output state at step $t$ conditioned on the dialogue context and partially decoded outputs.

Given the decoder state $\mathbf{m}_t$ and the concept flow memory $\mathbf{s}_N$, the following step is to select target concepts from $\mathcal{G}$. We first retrieve a group of subgraphs that corresponds to the concept set $f_N$ of the last utterance $u_N$ to prepare for the next round of concept transition. Here, each subgraph $g_{j}$ consists of a hit concept $c_{Nj}\in f_N$ and its concept neighbours. Formally, $g_j=\{(c^{head}_{j}, c^{tail}_{jk})\}^{N_{g_j}}_{k=1}$, where $c^{head}_j$ and $c^{tail}_{jk}$ represent head concept vertex and tail concept vertex, respectively. $N_{g_j}$ means the number of vertex pairs in $g_j$. Then, we employ a dynamic graph attention mechanism to calculate subgraph vectors $\mathbf{g}_j$ at each decoding step $t$ to fuse information of all concept neighbours:
\begin{align}
    \mathbf{g}_j &=\sum\nolimits^{N_{g_j}}_{k=1}\alpha^g_{jk}[\mathbf{e}^{head}_{j};\mathbf{e}^{tail}_{jk}],\nonumber \\
    \alpha^g_{jk} &= \frac{\mathrm{exp}(\beta^g_{jk})}{\sum^{N_{g_j}}_{l=1}\mathrm{exp}(\beta^g_{jl})}, \\
    \beta^g_{jk} &= (\mathbf{W}^g_q[\mathbf{m}_t;\mathbf{s}_N;\mathbf{e}^{head}_{j}])^{\top}\cdot(\mathbf{W}^g_k\mathbf{e}^{tail}_{jk}). \nonumber
\end{align}
$\mathbf{W}^g_q$, $\mathbf{W}^g_k$ are trainable parameters. $\mathbf{e}^{head}_j$, $\mathbf{e}^{tail}_{jk}$ are embeddings of head and tail concepts in $g_j$. Here, $\alpha^g_{jk}$ is the probability of choosing $c^{tail}_{jk}$ from all concept neighbours in $g_j$ at step $t$ conditioned on the dialogue context and the concept flow. We then compute the probability of choosing $g_j$ at step $t$ as a top-level concept selection, denoted as $\alpha^t_{j}$:
\begin{align}
    &\alpha^t_{j} = \frac{\mathrm{exp}(\beta^t_{j})}{\sum^m_{l=1}\mathrm{exp}(\beta^t_{l})},\nonumber \\
    &\beta^t_{j} = (\mathbf{W}^t_q[\mathbf{m}_t;\mathbf{s}_N])^{\top}\cdot(\mathbf{W}^t_k\mathbf{g}_{j}),
\end{align}
where $\mathbf{W}^t_q$ and $\mathbf{W}^t_k$ are trainable parameters. Finally, the selection probability of target concepts at step $t$ can be derived as:
\begin{equation}
    P(c_t|c_{1:t-1},D,F,\mathcal{G})=\alpha^t_j\alpha^g_{jk}.
\end{equation}

The multi-concept extractor has two stop conditions: 1) We add a special token $c^{stop}$ to the concept neighbour set of $g_j$\footnote{In this case, we make sure that $N_{g_j}>0$, where $g_j$ has at least one special vertex pair $(c^{head}_j,c^{stop})$.}. The extractor treats $c^{stop}$ as a legal candidate target, and the selection of $c^{stop}$ results in a stop action. 2) the number of target concepts exceeds $N_{max}$. Furthermore, for all the concepts extracted at step $k (k<t)$, we set their probabilities to 0 to avoid duplicate extraction.

\subsection{Concept-Guided Insertion Transformer}
After obtaining the target concept sequence $C$, the next step is to generate a response that covers $C$. General autoregressive approaches cannot ensure the success of introducing certain contents because they prefer to generate generic words \cite{mou2016sequence}. Given a substantially big language model, the problem might be alleviated but still cannot be completely solved. To address the issue, we use an Insertion Transformer \cite{stern2019insertion} to generate a response based on $C$, which ensures the appearance of target concepts. On the other hand, the explicit planned concepts can be regarded as a prompt or a signal to guide the generation process. Generation is accomplished by repeatedly making insertions into a sequence initialized by $C$ until a termination condition is met. At each decoding step $t$, the Insertion Transformer produces a joint distribution over the choice of words $w_t$ and all available insertion locations $l_t\in[0, |\hat{y}_{t-1}|]$ in the previously decoded response $\hat{y}_{t-1}$:
\begin{align}
    &\hat{y}_0 = C, \nonumber\\
    &\hat{\mathbf{E}}_{t} = [\mathbf{e}^w_k|w_k\in\hat{y}_{t}], \\
    &p(w_t,l_t|D, \hat{y}_{t-1}) =\mathrm{InsTF}( \mathbf{H}^{cls},\hat{\mathbf{E}}_{t-1}),\nonumber
\end{align}
where $\hat{\mathbf{E}_t}$ is the word embedding list of $\hat{y}_t$. Notably, $\hat{y}_t$ has multiple available insertion locations, and we can perform parallel decoding by applying insertions at multiple locations simultaneously. For more details of Insertion Transformer, please refer to Stern et al. \shortcite{stern2019insertion} due to the space limitation.
\subsection{Training and Loss Functions}
Given the list of ground truth concepts $C$ in the target response $y$, the concept extractor is trained as a usual sequence generation model to minimize the negative log likelihood (NLL) loss as follows:
\begin{equation}
    \mathcal{L}_C=\frac{1}{|C|}\sum\nolimits_{t=1}^{|C|}-\mathrm{log}p(c_t|c_{1:t-1},D,F,\mathcal{G}).
\end{equation}

To train the Insertion Transformer, we first sample a subsequence $\hat{y}$ containing all the target concepts from the target response $y$. Then, for each of the $k+1$ locations $l=0,1,...,k$ in $\hat{y}$, let $(w_{i_l},w_{i_l+1},...,w_{j_l})$ be the span of words from the target response yet to be produced at location $l$. The loss function is finally defined as follows:
\begin{equation}
    \mathcal{L}_R=\frac{1}{k+1}\sum^k_{l=0}\sum^{j_l}_{i=i_l}-\mathrm{log}p(w_i,l|D,\hat{y})\cdot w_l(i).
\end{equation}
Here $w_l(i)$ is a softmax weighting policy \cite{stern2019insertion} that performs a weighted sum of the negative log-likelihoods of the words in the span. It encourages the generator to produce the central words of the span for a faster decoding process.

\begin{table}[t!]
\small
\begin{center}
\setlength{\tabcolsep}{2mm}{
\begin{tabular}{lcc}
\toprule[1pt]
 &{\bf Persona} &{\bf Weibo} \\
\midrule
\# training pairs &101,935 &1,818,862\\
\# validating pairs & 5,602&9,187\\
\# testing pairs & 5,317&9,186\\
\# concept vertices in $\mathcal{G}$ &2,409 &4,000\\
\# transition edges in $\mathcal{G}$ &50,744 &74,362\\
\# concepts in each utterance &2.56 &1.61\\
\bottomrule[1pt]
\end{tabular}}
\end{center}
\caption{\label{tb:data}Statistics of the dialogue datasets and the constructed concept graphs. }
\end{table}

\section{Experimental Settings}
\begin{table*}[t!]
\small
\begin{center}
\setlength{\tabcolsep}{2mm}{
\begin{tabular}{|l|ccccc|ccccc|}
\hline
 \multirow{2}{*}{{\bf Model}}
 &\multicolumn{5}{c|}{{\bf Persona-Chat}} & \multicolumn{5}{c|}{{\bf Weibo}}\\

& BLEU& RG-1& RG-L& Dist-1& Dist-2 & BLEU& RG-1& RG-L& Dist-1&Dist-2\\
\hline
\multicolumn{11}{|c|}{{\em Without Concept Planning}} \\
\hline
Seq2seq+Att& .0325& .1698& .1691& .0127& .0402& .0106& .0790& .1004& .0160& .0520\\
Transformer& .0285& .1565& .1553& .0181&.0738 &.0287 & .1175&.1480 &.0186 &.0898 \\
HRED& .0332 & .1781& .1785& .0136& .0410& .0151& .0830& .1029&.0145 & .0730\\
ReCoSa& .0306& .1576& .1606& .0192&.0820 & .0232& .0858&.1069 & .0171&.0841 \\
\hline
\multicolumn{11}{|c|}{{\em With Concept Planning}} \\
\hline
Seq2BF& .0197& .1234& .1201& .0595&\bf .3058& .0157& .1506& .1434&\bf .0447& .1952\\
CCM& .0389& .1902& .1922& .0463& .1537&.0249 &.1883 & .1845& .0257& .1750\\
ConceptFlow & .0401&.2216 &.2154 & .0487& .1939&.0282 &.2133 & .2071& .0348& .1948\\
CG-nAR (ours)&\bf .0477&\bf.2611 &\bf.2502 &\bf.0626 &.2516 &\bf .0304 &\bf.2576 &\bf.2417 &.0401 &\bf.2809 \\
\hline
\end{tabular}}
\end{center}
\caption{\label{tb:auto_results} Results of automatic evaluation for CG-nAR and baseline methods, which are categorized into two groups: {\em with / without} concept planning. The best results are highlighted in bold.}
\end{table*}
\subsection{Datasets}
Experiments are conducted on two public open-domain dialogue datasets {\bf Persona-Chat} \cite{zhang2018personalizing} and {\bf Weibo} \cite{shang2015neural}. For Persona-Chat, the associated persona information is discarded so that the model can focus on the development of dialogues. Following previous works \cite{tang2019target,xu2020conversational}, we employ a rule-based method to automatically extract concept words of each utterance, which combines tf-idf and POS features for scoring word salience. After dataset cleaning, we re-split the Persona-Chat dataset into train/valid/test sets as done in Tang et al. \shortcite{tang2019target}, while the Weibo dataset is split in random. After constructing the graph of Persona-Chat, we randomly sample 100 concept vertices and 200 edges and ask three human annotators to evaluate their appropriateness. About 93\% vertices and 72\% edges are accepted by the annotators. For the graph of Weibo, we use the graph released by Xu et al. \shortcite{xu2020conversational}. Statistics of the two dialogue datasets along with the constructed graphs is shown in Table \ref{tb:data}.

\subsection{Comparison Methods}
We compare CG-nAR with two groups of baselines: general seq2seq models and concept-guided systems. General seq2seq models produce responses conditioned on the dialogue messages without concept planning, including: {\bf Seq2seq+Att} \cite{sutskever2014sequence}, a standard RNN model with attention mechanism; {\bf Transformer} \cite{vaswani2017attention}, a seq2seq model with a multi-head attention mechanism; {\bf HRED} \cite{serban2016building}, a hierarchical encoder-decoder framework to model context utterances; {\bf ReCoSa} \cite{zhang2019recosa}, a state-of-the-art model using the self-attention mechanism to measure the relevance of response and context. Concept-guided dialogue systems leverage concept information to control response generation, including: {\bf Seq2BF} \cite{mou2016sequence}, a non-left-to-right generation model that explicitly incorporates a keyword into the response; {\bf CCM} \cite{zhou2018commonsense}, a model that uses the graph attention mechanism to choose graph entities\footnote{The original CCM uses an external knowledge graph. Here we adapt it to our constructed concept graph for a fair comparison. The same strategy is applied to ConceptFlow.}, and introduces them into response implicitly by a copy mechanism; {\bf ConceptFlow} \cite{zhang2020grounded}, a state-of-the-art model that grounds each dialogue in the concept graph and traverses to distant concepts, which also generates concept words implicitly in an autoregressive manner; {\bf CG-nAR} (our model), a model that explicitly introduces multiple concepts into responses with non-autoregressive generation.

\subsection{Implementation Details}

We used VGAE \cite{kipf2016variational} to initialize the representation of concept vertices in the concept graph, and used Word2Vec \cite{mikolov2013distributed} to initialize word embeddings. The embedding size of vertices and words was set to 128 and 300, respectively. We employed Adam \cite{kingma2015adam} with learning rate 1e-3 to train the concept extractor and the Insertion Transformer. All Transformer blocks have 3 layers, 768 hidden units, 8 heads, and the hidden size for all feed-forward layers is 2,048. The hidden size of GRU cells is 768. At inference time, the multi-concept extractor produces concepts greedily, and the maximum number of allowed concepts $N_{max}$ was set to 5. For the Insertion Transformer, we used the configuration that achieved the best results reported in Stern et al. \shortcite{stern2019insertion}. The whole model was trained for 100,000 steps with 8,000 warm-up steps on a 3090 GPU. Checkpoints were saved and evaluated on the validation set every 2,000 steps. Checkpoints with the top performance were finally evaluated on the test set to report final results.

\section{Results and Analysis}
\subsection{Automatic Evaluation}
We adopt widely used {\em BLEU} \cite{papineni2002bleu} and {\em ROUGE} \cite{lin2004rouge} to measure the relevance between the generation and the ground-truth. We report averaged BLEU scores with 4-grams at most and ROUGE-1/L (RG-1/L) F-scores. To measure the diversity of generated responses, we report the ratio of distinct uni/bi-grams ({\em Dist-1/2}) in all generated responses \cite{li2016diversity}. 

Table \ref{tb:auto_results} shows 
the results of automatic evaluation for CG-nAR and baseline methods. All methods can be categorized into two groups: traditional seq2seq based generators and concept grounded methods. CG-nAR outperforms all other baselines significantly on BLEU and ROUGE scores (using Wilcoxon signed-rank test, $p$< 0.05), which manifests that the responses generated by CG-nAR match better with the ground-truth responses. This means CG-nAR can maintain the dialogue flow on-topic by the multi-concept planning mechanism. In terms of Dist-1/2 that measures the response diversity, all methods with concept planning can produce more diverse responses than those without, which indicates the problem of generic responses is alleviated by integrating concept information. Compared to the baselines with concept planning, CG-nAR has a better performance on response diversity. It verifies the effectiveness of our multi-concept planning module and the concept-guided non-autoregressive strategy, which can produce and combine multiple context-related concepts to compose diverse responses and keep concept words in the output response in an explicit manner.

\begin{table}[t]
\small
\begin{center}
\setlength{\tabcolsep}{2mm}{
\begin{tabular}{|lcc|}
\hline
{\bf Model} &\bf App.&\bf Inf.\\
\hline
\multicolumn{3}{|l|}{{\em Persona-Chat}} \\
CG-nAR vs. ReCoSa& 72.5\%$^\dagger$&59.3\%$^\dagger$ \\
CG-nAR vs. Seq2BF& 63.0\%$^\dagger$& 53.1\%\\
CG-nAR vs. CCM& 60.5\%$^\dagger$& 55.0\% \\
CG-nAR vs. ConceptFlow& 58.2\%$^\dagger$& 54.4\% \\
\hline
\multicolumn{3}{|l|}{{\em Weibo}} \\
CG-nAR vs. ReCoSa& 78.2\%$^\dagger$&61.9\%$^\dagger$ \\
CG-nAR vs. Seq2BF& 65.5\%$^\dagger$& 52.9\%\\
CG-nAR vs. CCM& 61.4\%$^\dagger$& 54.0\% \\
CG-nAR vs. ConceptFlow& 61.1\%$^\dagger$& 54.7\% \\
\hline
\end{tabular}}
\end{center}
\caption{\label{tb:human_results} Results of manual evaluation with {\em appropriateness} (App.) and {\em informativeness} (Inf.). The score is the percentage of times CG-nAR is chosen as the better in pairwise comparison with its competitor. Results marked with $\dagger$ are significant (using sign test, $p$<0.05).}
\end{table}

\subsection{Manual Evaluation}
Considering automatic metrics may not suitably reflect the content to be evaluated, we further performed manual evaluation following previous works \cite{zhou2018commonsense,wu2020topicka}. Specifically, we randomly sampled 200 testing pairs from each test set and employed three annotators with professional background knowledge to evaluate the responses. Given a dialogue message, annotators were required to conduct pair-wise comparison between the response generated by CG-nAR and the one by a baseline (1,600 comparisons with four baselines on two datasets in total). For each comparison, annotators decided which response is better in terms of {\em appropriateness} (the model's ability to produce a fluent, coherent, and context-relevant response) and {\em informativeness} (if the response provides diverse information). For appropriateness, the percentage of pairs that at least 2 annotators gave the same judge (2/3 agreement) is 95.8\%, and the percentage for 3/3 agreement is 62.7\%. For informativeness, the at least 2/3 agreement is 89.0\% and 3/3 agreement is 56.2\%.

We compare CG-nAR against four baselines on Persona-Chat and Weibo (see Table \ref{tb:human_results}). The score represents the percentage of times CG-nAR is chosen as the better in pair-wise comparisons. For appropriateness, CG-nAR significantly outperforms all other baselines on two datasets (using sign test, $p$ < 0.05). It means that CG-nAR can generate more context-relevant and coherent responses accepted by annotators, which validates the effectiveness of our multi-concept planning module. In terms of informativeness, the percentages that CG-nAR wins ReCoSa are noticeably higher than those against other baselines. It indicates that systems with a concept planning mechanism can produce more informative responses by content introducing. 

\begin{table}[t]
\small
\begin{center}
\setlength{\tabcolsep}{2mm}{
\begin{tabular}{|lcccc|}
\hline
{\bf Model} &\bf P&\bf R & \bf F1 &\bf Num.\\
\hline
\multicolumn{5}{|l|}{{\em Persona-Chat}} \\
ReCoSa&.0137 & .0077& .0099& 1.29\\
Seq2BF& .0280& .0189& .0226& 1.55\\
CCM& .2406&.1853 & .2094&1.34 \\
ConceptFlow& .3580&.4041 & .3797& 1.50\\
CG-nAR&\bf.5330 & \bf.5029&\bf .5175&2.17\\
\hline
\multicolumn{5}{|l|}{{\em Weibo}} \\
ReCoSa& .0643&.0611 & .0626&1.53 \\
Seq2BF& .1685&.1657 & .1671& 1.58\\
CCM&.2514 & .2059&.2264 & 1.61\\
ConceptFlow& .3859& .4177&.4012 & 1.78\\
CG-nAR&\bf .5119&\bf.6455 &\bf.5710 &2.03\\
\hline
\end{tabular}}
\end{center}
\caption{\label{tb:concept_results}Results of {\bf Concept-P/R/F1} that compare the concepts in output responses with those in ground-truth ones. {\bf Num.} denotes the average number of concepts predicted in output responses. }
\end{table}

\subsection{Analysis of Multi-Concept Planning}

\begin{table*}[t!]
\small
\begin{center}
\setlength{\tabcolsep}{2mm}{
\begin{tabular}{|l|cccccccc|}
\hline
 {\bf Model}& \bf BLEU& \bf RG-1& \bf RG-L&\bf Dist-1& \bf Dist-2 & \bf Concept-P&\bf Concept-R& \bf Concept-F1 \\
\hline
CG-nAR&\bf .0477&\bf .2611&\bf .2502&\bf .0626& .2516& .5330& \bf.5029&\bf.5175  \\
\ \ w/o. concept planning& .0217&.1504 &.1568 &.0350 &.1884 &.2711 &.2507 & .2605\\
\ \ w/o. concept flow encoder& .0344 &.2283& .2232&.0583 &.1938 &\bf.5778 &.3696 &.4508 \\
\ \ w/o. hierarchical encoder& .0327&.2130 & .2009& .0468&\bf.2650 &.3242 & .4250&.3678 \\
\ \ w. \ \ \  gated controller (AR)& .0405&.2399 & .2320& .0423&.1893 &.3926 & .4177&.4048 \\
\hline
\end{tabular}}
\end{center}
\caption{\label{tb:ablation} Ablation study using automatic metrics on the Persona-Chat dataset. The best results are highlighted.}
\end{table*}

\begin{table}[t]
\small
\begin{center}
\setlength{\tabcolsep}{2mm}{
\begin{tabular}{|lccc|}
\hline
{\bf Model} & \bf Param.& {\bf total time} (sec)&\bf words/sec \\
\hline
Transformer& 127.3M &71.36 &672.1\\
ReCoSa& 143.8M& 65.27& 766.9\\
CG-nAR&172.1M & 25.99& 1131.1\\
\hline
\end{tabular}}
\end{center}
\caption{\label{tb:speed_results} Inference speed on the Persona-Chat test set. {\bf Param.} denotes the number of parameters.}
\end{table}

\begin{table*}[t]
\small
\begin{center}
\setlength{\tabcolsep}{2mm}{
\begin{tabular}{|l|l|l|}
\hline
 \multirow{6}{*}{Context}& A: What do you do for work?
&A: Hi. How are you doing?\\
&B: No work just the {\color{blue}hits} that's all I need.&
B: I'm good, just {\color{blue}finished practicing} the \\
 &A: I see, I am a {\color{blue}volunteer} at my {\color{blue}local}& \quad {\color{blue}guitar}. You?\\
 &  \quad {\color{blue}animal shelter}. & B: Do you do that for a {\color{blue}living}?\\
 &B: Well this good work. I am a {\color{blue}veteran} and I
&A: No, just a {\color{blue}hobby} because {\color{blue}country music}
\\
& \quad  have a {\color{blue}garden}.& \quad is my {\color{blue}favorite}.\\
\hline
\multirow{2}{*}{Ground Truth}& Thank you for your {\color{red}service}! Do you {\color{red}grow} &\multirow{2}{*} {Who is your {\color{red}favorite} {\color{red}singer}?}\\
&\quad any {\color{red}vegetables}? & \\
\hline
ReCoSa& That's cool. Do you have any pets?&I'm a teacher. I've a dog.\\
Seq2BF &Cool! I work part time at an {\color{red}animal shelter}. &I like {\color{red}music}. I love my job.\\
CCM&I'm a teacher. What do you do?&Do you have any {\color{red}hobbies}?\\
ConceptFlow & I enjoy to eat {\color{red}organic foods}. &I love {\color{red}music}. What are your {\color{red}hobbies}?\\
CG-nAR& Do you {\color{red}grow vegetable} for a {\color{red}living}?& That's cool. Who is your {\color{red}favorite singer}?\\
\hline
\end{tabular}}
\end{center}
\caption{\label{tb:case} Dialogue cases with output responses from different systems. Words in {\color{blue}Blue} are the observed concepts in the dialogue flow. Words in {\color{red}Red} represent the context-associated concepts in the output response. }
\end{table*}

To validate if the multi-concept planning module has the ability to extract context-relevant concepts and form a coherent dialogue, we calculate the precision, recall, and F1 score of predicted concepts against golden ones in responses (Concept-P/R/F1). We also record the average number of predicted concepts to measure the model’s ability to introduce multiple concepts. From Table \ref{tb:concept_results} we can observe that CG-nAR achieves a higher recall and F1 score against all baselines by a large margin, especially for ReCoSa and Seq2BF. It probes that our concept planning module can successfully extract more concepts relevant to the dialogue. This is also reflected in the number of predicted concepts, where CG-nAR produces more concept words than those methods with autoregressive generators, e.g., CCM and ConceptFlow. It indicates that the concept-guided generator can effectively keep the concept information in output responses using a non-autoregressive generation mechanism. 

\subsection{Ablation Study}
We perform ablation studies to validate the effectiveness of each part of CG-nAR. Table \ref{tb:ablation} shows the results. One of the variants is a vanilla Insertion Transformer where the concept planning module is removed. The model performance unsurprisingly degrades by a large margin, because the model might produce generic responses without concept planning. After removing the concept flow encoder, the information of historical concept transitions is missing, which also leads to a performance drop. We  further replace the hierarchical dialogue encoder with a vanilla Transformer encoder, the performance drop shown in Table \ref{tb:ablation} indicates 
that it is necessary to capture the context dependency information when performing dialogue modeling. To probe the effectiveness of the concept-guided non-autoregressive strategy, we replace the Insertion Transformer with a universal Transformer framework equipped with a gated controller as done in Zhang et al. \shortcite{zhang2020grounded}, where the generation probabilities are calculated over the word vocabulary and the set of selected concept words. Table \ref{tb:ablation} shows that with the autoregressive decoding strategy, the performance drop is significant. A possible explanation is that the appearance of some key concepts cannot be guaranteed by such an implicit concept-oriented generator, especially when the generator encounters concepts that are not frequently seen in the training set.

\subsection{Speed Comparison}

Our concept-guided non-autoregressive generation model shows not only the superiority on response quality, but also gives a significant speed-up at test time over the methods equipped with autoregressive generators. The results of speed comparison is shown in Table \ref{tb:speed_results}. For a fair comparison, we choose the baselines with a Transformer encoder-decoder framework, since our customized Insertion Transformer uses the same model components. The main advantage of the insertion-based generator at inference time is that we can predict words at different insertion locations  simultaneously. From Table \ref{tb:speed_results} we can see that CG-nAR achieves substantially test-time speed-up compared to the two autoregressive generators (up to 2.7x in total time and 1.6x in word generation rate) even when CG-nAR has more parameters\footnote{Here we test the autoregressive baselines with a beam size of 3 (used for their best scores). Without beam-search, they have significantly worse results, so we do not compare speed-ups with that version.}.

\subsection{Case Study}
To compare different models intuitively, we show two dialogue cases of the Persona-Chat dataset with output responses in Table \ref{tb:case}. We observe that CG-nAR can successfully output context-associated concepts, e.g., {\em grow vegetable} that is related to {\em garden}, and {\em singer} that is related to {\em country music}. Compared to other baselines, CG-nAR produces a response that is more coherent and relevant to the dialogue context, and shows a more natural transition of concepts, which again proves the effectiveness of our concept-guided non-AR strategy for controllable dialogue generation.

\section{Conclusion}
In this work, we propose a novel concept-guided non-autoregressive approach for open-domain dialogue generation. It consists of a multi-concept planning module that selects multiple context-relevant concepts to facilitate a coherent dialogue, and a customized Insertion Transformer that produces a response based on the selected concepts to control the generation process. The experimental results show that our method can not only produce high-quality responses, but can also significantly speed up the inference time. 

\section*{Acknowledgements}
The authors wish to thank the anonymous reviewers for their helpful comments. This work was partially funded by China National Key R\&D Program (No. 2017YFB1002104), National Natural Science Foundation of China (No. 61976056, 62076069), Shanghai Municipal Science and Technology Major Project (No.2021SHZDZX0103).

\bibliography{anthology,custom}
\bibliographystyle{acl_natbib}

\end{document}